\documentclass[wcp]{jmlr}

\usepackage{longtable}

\usepackage{booktabs}
\usepackage{siunitx}
\usepackage{fontawesome}
\usepackage{mathtools}
\usepackage{wrapfig}
\usepackage{algorithm}
\usepackage{algorithmic}

\makeatletter
\let\Ginclude@graphics\@org@Ginclude@graphics 
\makeatother

\newcommand{\matr}[1]{\mathbf{#1}}
\newcommand{\norm}[1]{\left\lVert#1\right\rVert}

\DeclarePairedDelimiter\abs{\lvert}{\rvert}

\DeclarePairedDelimiter\floor{\lfloor}{\rfloor}
\newcommand{\defeq}{\vcentcolon=}

\jmlrvolume{157}
\jmlryear{2021}
\jmlrworkshop{ACML 2021}

\title[Sinusoidal Flow]{Sinusoidal Flow: A Fast Invertible Autoregressive Flow}

  \author{\Name{Yumou Wei} \Email{yumouwei@umich.edu}\\
  \addr University of Michigan --- Ann Arbor, MI USA
 }

\editors{Vineeth N Balasubramanian and Ivor Tsang}

\begin{document}

\maketitle

\begin{abstract}
Normalising flows offer a flexible way of modelling continuous probability distributions. We consider expressiveness, fast inversion and exact Jacobian determinant as three desirable properties a normalising flow should possess. However, few flow models have been able to strike a good balance among all these properties. Realising that the integral of a convex sum of sinusoidal functions squared leads to a bijective residual transformation, we propose Sinusoidal Flow, a new type of normalising flows that inherits the expressive power and triangular Jacobian from fully autoregressive flows while guaranteed by Banach fixed-point theorem to remain fast invertible and thereby obviate the need for sequential inversion typically required in fully autoregressive flows. Experiments show that our Sinusoidal Flow is not only able to model complex distributions, but can also be reliably inverted to generate realistic-looking samples even with many layers of transformations stacked.
\end{abstract}

\section{Introduction}

Estimating arbitrarily complex probability distributions from data is a critical task in machine learning, as an accurate specification of the underlying probabilistic model allows us to perform useful statistical computations such as generating samples or making inferences. Among all the tools available, normalising flows~\citep{https://doi.org/10.1002/cpa.21423,pmlr-v37-rezende15} appear as a highly flexible method for modelling continuous probability distributions by constructing a differentiable, bijective transformation $T: \mathcal{Z} \subseteq \mathbb{R}^{D} \mapsto \mathbb{R}^{D}$ that transforms a random vector $\matr{z} \in \mathbb{R}^{D}$, whose distribution is known and supported on $\mathcal{Z}$, to the random vector $\matr{x} \in \mathbb{R}^{D}$ whose distribution is to be estimated. The distribution of $\matr{z}$ is usually chosen to be a simple one, such as the standard multivariate Gaussian distribution:
\begin{equation}
    \matr{x} = T\left(\matr{z}\right), \; \mathrm{where} \; \matr{z} \sim p_{\matr{z}}\left(\matr{z}\right) = \mathcal{N}\left(\matr{z} \mid \matr{0}, \matr{I}\right)
\end{equation}

The change-of-variable formula for probability density functions~\citep{CaseBerg:01} then allows the unknown density $p_{\matr{x}}$ to be specified in terms of the known density $p_{\matr{z}}$:
\begin{equation}\label{eqn: formulation}
    p_{\matr{x}}\left(\matr{x}\right) = p_{\matr{z}}\left(T^{-1}\left(\matr{x}\right)\right) 
    \cdot 
    \abs{\mathrm{det}\, \matr{J}_{T^{-1}}\left(\matr{x}\right)}
\end{equation}
where $\matr{J}_{T^{-1}}\left(\matr{x}\right)$ is the $D \times D$ Jacobian matrix of the inverse transformation $T^{-1}$. Depending on how $T^{-1}$ is constructed, normalising flows can be broadly categorised into discrete-time flows and continuous-time flows~\citep{papamakarios2019normalizing}. 

Discrete-time flows feature a decomposition of $T^{-1}$ into a finite number of simpler transformations:
\begin{align}
    T^{-1} &= T^{-1}_{L} \circ T^{-1}_{L - 1} \circ \cdots \circ T^{-1}_{1} \label{eqn: decomp} \\
    \mathrm{det}\, \matr{J}_{T^{-1}} &= \mathrm{det}\, \matr{J}_{T^{-1}_{L}} \times \mathrm{det}\, \matr{J}_{T^{-1}_{L - 1}} \times \cdots \times \mathrm{det}\, \matr{J}_{T^{-1}_{1}}
\end{align}
where each constituent transformation $T_{l}^{-1}$ can be implemented as a neural network layer and the calculation of the overall Jacobian determinant can be decentralised to each layer. Depending on how each constituent transformation is constructed, discrete-time flows can be further divided into fully autoregressive, partially autoregressive and residual flows~\citep{papamakarios2019normalizing}, which we describe in Section~\ref{sec: related_work}, along with continuous-time flows. 

We think the following three properties, among others, desirable for normalising flows. 

\textbf{Expressiveness:} how well a normalising flow can represent an arbitrary probability distribution. The expressive power of a normalising flow largely depends on how the transformation $T^{-1}$ (or equivalently, $T$) is constructed. Models that advocate strong dependency among the components of $\matr{x}$, such as fully autoregressive flows and residual flows, appear to exhibit strong expressive power as well. In fact, fully autoregressive flows are shown to be universal approximators under mild assumptions about $p_{\matr{x}}$~\citep{papamakarios2019normalizing}, in that they can accurately represent any probability distributions given a transformation $T^{-1}$ with sufficient capacity. 

\textbf{Fast inversion:} whether a normalising flow can be inverted computationally efficiently. The formulation of normalising flows in Equation~\ref{eqn: formulation} is well-suited for density estimation; however, to generate samples from $p_{\matr{x}}$ we would need to invert $T^{-1}$ to obtain $T$. Certainly, we can always invert any differentiable, theoretically invertible functions using a common root-finding algorithm, such as bisection or Newton's method with a quadratic convergence rate, but we are interested in whether we can build normalising flows with an analytic inverse or with an even faster numerical inversion procedure that is also less sensitive to the choice of starting points. Due to the autoregressive dependency imposed on $\matr{x}$, fully autoregressive flows in general have to be inverted sequentially, one component at a time; moreover, depending on how each constituent transformation $T_{l}^{-1}$ is implemented component-wise, there may or may not exist a faster, more reliable inversion procedure than the Newton's method. Partially autoregressive flows exchange some autoregressiveness for analytic invertibility at the cost of reduced expressive power. However, faster numerical inversion procedures do exist for residual flows and continuous-time flows, which we describe in Section~\ref{sec: related_work}. 

\textbf{Exact Jacobian determinant:} whether the $\mathrm{det}\, \matr{J}_{T^{-1}}\left(\matr{x}\right)$ term in Equation~\ref{eqn: formulation} can be calculated exactly and efficiently. We can always compute the full Jacobian matrix of any neural network layer and its determinant at a cost of $\mathcal{O}\left(D^{3}\right)$, which is however intractable for very large $D$ as often seen in images or videos. So one of the focuses of previous work has been designing special architectures for the transformation $T^{-1}$ so that the Jacobian determinant can be either calculated exactly or estimated efficiently, both at a cost no more than $\mathcal{O}\left(D\right)$. Autoregressive flows feature a triangular Jacobian matrix at each layer whose determinant can be calculated exactly and efficiently by design. However, the Jacobian determinant for residual flows and continuous-time flows in general needs to be estimated or computed exactly at a cost slightly higher than $\mathcal{O}\left(D\right)$. 

Table~\ref{tbl: properties} summarises our best evaluation of some selected autoregressive, residual and continuous-time flows against the three desirable properties we propose. To the best of our knowledge, few normalising flows seem to have claimed excellence in all three dimensions. Fully autoregressive flows would have so if only they were as efficiently invertible as residual flows, whereas residual flows would have so if only they were made autoregressive to allow exact Jacobian determinant calculation. As we show in Section~\ref{sec: sinusoidal}, our proposed Sinusoidal Flow is simultaneously a fully autoregressive flow and a residual flow, which strikes a good balance among all the three desirable properties.

\begin{table*}[t]
\caption{Evaluation of selected normalising flows against the three desirable properties we propose. The ratings for expressiveness represent our best subjective assessment on each method primarily based on the objective measures (if available) shown in Table~\ref{tbl: uci_results}. We understand that different normalising flows have different primary use cases. Normalising flows with convolutional architectures such as Glow~\citep{NEURIPS2018_d139db6a}, MaCow~\citep{NEURIPS2019_20c86a62} and  MintNet~\citep{NEURIPS2019_70a32110} are better at modelling images. The ratings would have been different should we refer to a different objective measure. }
\vskip -0.1in
\label{tbl: properties}
\begin{center}
\begin{sc}
\scalebox{0.72}{
\begin{tabular}{lcccccr}
\toprule
Method & Expressiveness & Fast Inversion & Exact Jacobian Det. \\
\midrule
\textbf{Fully Autoregressive Flow} \\
\hspace{1em} MAF~\citep{NIPS2017_6c1da886}     & \faStar\;\faStar\;\faStarO & $\times$ & $\surd$ \\
\hspace{1em} NAF~\citep{pmlr-v80-huang18d}      & \faStar\;\faStar\;\faStar & $\times$ & $\surd$       \\
\hspace{1em} B-NAF~\citep{pmlr-v115-de-cao20a}   & \faStar\;\faStar\;\faStar & $\times$ & $\surd$   \\
\hspace{1em} SOS~\citep{pmlr-v97-jaini19a}   & \faStar\;\faStar\;\faStar & $\times$ & $\surd$  \\
\hspace{1em} UMNN-MAF~\citep{NEURIPS2019_2a084e55}   & \faStar\;\faStar\;\faStar & $\times$ & $\surd$ \\
\hspace{1em} 
RQ-NSF(AR)~\citep{NEURIPS2019_7ac71d43}   & \faStar\;\faStar\;\faStar & $\times$ & $\surd$ \\
\midrule
\textbf{Partially Autoregressive Flow} \\
\hspace{1em} RealNVP~\citep{DBLP:conf/iclr/DinhSB17}    & \faStar\;\faStarO\;\faStarO & $\surd$ & $\surd$ \\
\hspace{1em} Glow~\citep{NEURIPS2018_d139db6a} & \faStar\;\faStarO\;\faStarO & $\surd$ & $\surd$\\
\hspace{1em} RQ-NSF(C)~\citep{NEURIPS2019_7ac71d43} & \faStar\;\faStar\;\faStar & $\surd$ & $\surd$\\
\midrule
\textbf{Residual Flow} \\
\hspace{1em} i-ResNet~\citep{pmlr-v97-behrmann19a} & \faStar\;\faStar\;\faStar & $\surd$ & $\times$ \\
\midrule
\textbf{Continuous-time Flow} \\
\hspace{1em}  FFJORD~\citep{grathwohl2018scalable} & \faStar\;\faStar\;\faStarO & $\surd$ & $\times$ \\
\midrule
\textbf{Sinusoidal Flow (Ours)}   & \faStar\;\faStar\;\faStar & $\surd$ & $\surd$  \\
\bottomrule
\end{tabular}
}
\end{sc}
\end{center}
\vskip -0.2in
\end{table*}

\section{Related Work}\label{sec: related_work}

In this section, we review and analyse previous architectures of normalising flows related to our work, particularly with respect to the three desirable properties we propose. 

\subsection{Autoregressive Flows}\label{sec: autoregressive}

Autoregressive flows are usually specified in terms of a ``transformer'' $\tau$ acting on each component of the input vector and a ``conditioner'' $c_{i}$ that computes the parameters of the transformer autoregressively based on the input. In other words, each constituent transformation $T_{l}^{-1}$ in Equation~\ref{eqn: decomp} consists of the following transformation for each component of the input $\matr{z}^{l - 1}$ and the corresponding output $\matr{z}^{l}$ at layer $l$, if we let $\matr{x} = \matr{z}^{0}$ and $\matr{z} = \matr{z}^{L}$:
\begin{equation}\label{eqn: auto_framework}
    z_{i}^{l} = \tau(z_{i}^{l - 1}; \matr{h}_{i}), \; \text{where} \; \matr{h}_{i} = c_{i}(\matr{z}_{<i}^{l - 1})
\end{equation}
The conditioner $c_{i}$ for the $i$-th input component takes all the components preceding the $i$-th and computes the parameters $\matr{h}_{i}$ for the transformer $\tau$. Sometimes though, it may be desirable for the conditioner $c_{i}$ to output parameters that are $\textit{independent}$ from the input, in which case we shall call the conditioner an \textit{independent conditioner}. 

\subsubsection{The Transformers}

The transformer $\tau$ can assume many functional forms, so long as it is differentiable and invertible. Usually it is designed as a strictly monotonic function, since strict monotoni\-city implies invertibility.  Popular choices include affine transformers~\citep{dinh2015nice,DBLP:conf/iclr/DinhSB17,NIPS2016_ddeebdee,NIPS2017_6c1da886,NEURIPS2018_d139db6a}, neural transformers~\citep{pmlr-v80-huang18d,pmlr-v97-ho19a,pmlr-v115-de-cao20a}, spline-based transformers~\citep{10.1145/3341156,NEURIPS2019_7ac71d43} and integration-based transformers~\citep{pmlr-v97-jaini19a,NEURIPS2019_2a084e55}. 

The design of the transformers greatly affects the availability of efficient inversion procedures. More complex transformers are usually accompanied by a less efficient inversion procedure, but less complex transformers usually come at a cost of reduced expressive power. For example, the neural network backbone in NAF~\citep{pmlr-v80-huang18d}, B-NAF~\citep{pmlr-v115-de-cao20a} and UMNN-MAF~\citep{NEURIPS2019_2a084e55} is shown to be very powerful, yet no faster inversion procedures other than the general bisection or Newton's method are proposed by the authors. On the other hand, IAF~\citep{NIPS2016_ddeebdee} and MAF~\citep{NIPS2017_6c1da886} employ trivially invertible affine transformers but are less expressive than those using neural transformers.

\subsubsection{The Conditioners}

The fundamental impediment to autoregressive flows' efficient inversion lies in their characteristic autoregressive dependency, which is directly related to the design of the \textit{conditioners}. 

\textbf{Fully autoregressive flows} are equipped with conditioners that impose the fullest possible autoregressive dependency on the parameters $\matr{h}_{i}$ of each transformer. Each $\matr{h}_{i}$ depends on \emph{all} preceding input components $\matr{z}_{<i}^{l - 1}$ and each input component $z_{i}^{l - 1}$ is non-trivially transformed by a conditional transformer $\tau\left(\cdot\; ; \matr{h}_{i}\right)$. Popular choices for the conditioner include masked feed-forward networks from MADE~\citep{pmlr-v37-germain15}, as used in 
IAF~\citep{NIPS2016_ddeebdee}, MAF~\citep{NIPS2017_6c1da886}, B-NAF~\citep{pmlr-v115-de-cao20a}, SOS~\citep{pmlr-v97-jaini19a}, RQ-NSF(AR)~\citep{NEURIPS2019_7ac71d43} and UMNN-MAF~\citep{NEURIPS2019_2a084e55}, and masked convolutional networks, as used in MaCow~\citep{NEURIPS2019_20c86a62} and  MintNet~\citep{NEURIPS2019_70a32110}. The fully autoregressive dependency necessitates a sequential inversion: before inverting ${z}_{i}^{l}$ at layer $l$, all $\matr{z}_{<i}^{l - 1}$ must first be recursively recovered from $\matr{z}_{<i}^{l}$ so that the conditioner $c_{i}$ can return the parameters of $\tau$, followed by a possibly non-trivial inversion procedure if $\tau$ is not so simple as an affine transformer. 

\textbf{Partially autoregressive flows} offer an analytic inverse in place of sequential inversion by imposing autoregressive dependency on only part of the input. Specifically, at each layer, half of the input is left unchanged and is used by a shared conditioner to produce the parameters of the individual transformers for the other half of the input:
\begin{equation}
\begin{cases}
    z^{l}_{i} = z^{l - 1}_{i} & \forall i\leq\floor{D/2} \\
    z^{l}_{i} = \tau(z^{l-1}_{i}; \matr{h}_{i}) & \forall i > \floor{D/2}, \; \text{where} \; \left[\matr{h}_{\floor{D/2} + 1}, \dots, \matr{h}_{D} \right] = c\left(\matr{z}^{l-1}_{i\leq\floor{D/2}}\right)
\end{cases}
\end{equation}
This special architecture is also known as coupling layers. If the transformer $\tau$ is analytically invertible, then the entire flow is analytically invertible, as in NICE~\citep{dinh2015nice}, RealNVP~\citep{DBLP:conf/iclr/DinhSB17},  Glow~\citep{NEURIPS2018_d139db6a} and RQ-NSF(C)~\citep{NEURIPS2019_7ac71d43}. However, the analytic invertibility comes at a cost of reduced expressive power in general compared to their fully autoregressive counterparts; nevertheless, the use of multi-scale convolutional architectures makes them more suitable for modelling images. 

In short, while inducing strong expressive power, fully autoregressive dependency greatly limits the availability of efficient inversion procedures for fully autoregressive flows. Partially autoregressive flows bypass the limitation at the expense of expressiveness. Both of them, however, feature a triangular Jacobian at each layer whose determinant can be calculated exactly and efficiently, thanks to the autoregressive dependency. 

\subsection{Residual Flows} \label{sec: residual}

In residual flows, each constituent transformation $T_{l}^{-1}$ in Equation~\ref{eqn: decomp} closely resembles a residual network~\citep{7780459}:
\begin{equation}
    \matr{z}^{l} = T_{l}^{-1}(\matr{z}^{l - 1}) = \matr{z}^{l - 1} + f(\matr{z}^{l - 1}; \phi)
\end{equation}
which can be made invertible with a special choice of  $f$. 

One possible design is constraining $f$ to be Lipschitz continuous with a Lipschitz constant $0 \leq M < 1$, so that $f$ becomes a contraction mapping. As noted in~\citet{pmlr-v97-behrmann19a}, the Banach fixed-point theorem guarantees that every contraction mapping has a unique fixed point, which gives rise to a very efficient inversion procedure for contractive residual flows. As~\citet{papamakarios2019normalizing} shows, if $f$ is a contraction mapping, then the function $F\left(\matr{\hat{z}}\right) = \matr{z}^{l} - f\left(\matr{\hat{z}}; \phi\right)$
is also a contraction mapping with a unique fixed point $\matr{z}^{\ast}$:
\begin{equation}\label{eqn: fixed_point}
    F(\matr{z}^{\ast}) = \matr{z}^{l} - f\left(\matr{z}^{\ast}; \phi\right) = \matr{z}^{\ast}
\end{equation}
which implies $\matr{z}^{l} = T_{l}^{-1}\left(\matr{z}^{\ast}\right)$ after a rearrangement of the terms. So the unique fixed point $\matr{z}^{\ast}$ of $F$ is also the unique pre-image of $\matr{z}^{l}$ under $T_{l}^{-1}$, which can be found by a simple fixed-point iteration algorithm proposed by~\citet{pmlr-v97-behrmann19a}. The algorithm is guaranteed to converge with \textit{any} starting points, at a rate \textit{exponential} in the number of iterations. 

The contractive residual transformation expands the possibilities for a more efficient inversion procedure. As we shall see in Section~\ref{sec: sinusoidal}, our proposed Sinusoidal Flow leverages exactly this special structure to allow fast inversion, even though being fully autoregressive in nature. On the other hand, the contractive residual transformation also imposes a dense dependency structure on the input $\matr{z}^{l - 1}$ and output $\matr{z}^{l}$, as evidenced by its dense Jacobian matrix whose determinant is in general not attainable at a cost no more than $\mathcal{O}\left(D\right)$ and usually requires a numerical estimator. Therefore, residual flows that rely on matrix identities to simplify determinant calculation, such as planar flows~\citep{pmlr-v37-rezende15} and Sylvester flows~\citep{berg2019sylvester}, are proposed. However, their expressive power is limited by the special structures introduced to make the identities applicable. 

In short, residual flows that employ a contractive residual transformation can be fast and reliably inverted, and usually possess expressive power comparable to that of autoregressive flows. However, the Jacobian determinant is not as readily available as with autoregressive flows due to their dense Jacobians in general. There exist other residual flows with more efficient Jacobian determinant calculation but reduced expressive power. 

\subsection{Continuous-time Flows}
Continuous-time flows are substantively different from all the normalising flows that have been reviewed so far and the Sinusoidal Flow we propose. They are characterised by ODEs rather than a finite composition of simpler transformations as in Equation~\ref{eqn: decomp}:
\begin{equation}
    \frac{d\matr{z}_{t}}{d t} = g\left(t, \matr{z}_{t}; \phi\right)
\end{equation}
where $g$ is a neural network parameterised by $\phi$ that is uniformly
Lipschitz continuous in $\matr{z}_{t}$ and continuous in $t$~\citep{NEURIPS2018_69386f6b, papamakarios2019normalizing}. If we let $\matr{x} = \matr{z}_{t_{0}}$ and $\matr{z} = \matr{z}_{t_{1}}$, the transformation $T^{-1}$ is given by an integration over time:
\begin{equation}
    T^{-1}\left(\matr{x}\right) = \matr{x} - \int_{t_{0}}^{t_{1}} g\left(t, \matr{z}_{t}; \phi\right) dt
\end{equation}
which can be inverted at the same cost as evaluating $T^{-1}$ by running the same integration backward in time. 

Continuous-time flows can be inverted as efficiently as their forward evaluation and have demonstrated strong expressive power in practice. However, methods like FFJORD~\citep{grathwohl2018scalable} generally need a stochastic estimator for the log-determinant of the Jacobian to avoid the very high cost of computing it exactly. 

\section{Sinusoidal Flow}\label{sec: sinusoidal}

In this section, we introduce our Sinusoidal Flow and demonstrate that it strikes a good balance among the three desirable properties we put forward, namely, expressiveness, fast inversion and exact Jacobian determinant. From Section~\ref{sec: autoregressive} we know that fully autoregressive flows already exhibit strong expressive power and feature a simple triangular Jacobian at each layer, offering an excellent basis for building new normalising flows that excel at all three dimensions. Therefore, we formulate our Sinusoidal Flow as a fully autoregressive flow composed of transformers and conditioners, which we shall first describe.   

\subsection{Sinusoidal Transformers}

It is an easily provable fact that the anti-derivatives of a positive-valued continuous function are monotonically increasing and hence invertible, which suggests a viable way of constructing a differentiable and invertible component-wise transformer $\tau$ as follows:
\begin{equation}\label{eqn: integral}
    z^{l}_{i} = \tau(z^{l - 1}_{i};\matr{h}_{i}, d_{i}) = \int_{0}^{z^{l - 1}_{i}} f(t;\matr{h}_{i}) dt + d_{i} , \; \text{where} \; \left[\matr{h}_{i}, d_{i}\right] = c_{i}\left(\matr{z}_{<i}^{l - 1}\right)
\end{equation}
where $f(t) > 0$ for all $t \in \mathbb{R}$ and $\left[\matr{h}_{i}, d_{i}\right]$ are the parameters of the transformer produced by a (possibly independent) conditioner $c_{i}$ as in autoregressive flows. In fact, some existing work has already explored this integration-based formulation; for example, UMNN-MAF~\citep{NEURIPS2019_2a084e55} implements $f$ as a positively constrained neural network, which however entails a numerical solver to approximate the integral. One case where the integral can be solved exactly is when $f$ is taken to be a sum of squares of polynomials, as proposed in SOS~\citep{pmlr-v97-jaini19a}. Inspired by their work, we propose taking $f$ to be a \textit{convex} sum of squares of \textit{sinusoidals}:
\begin{equation}
    f(t;\matr{h}_{i}) = 2\sum_{k = 1}^{K} w_{k}\sin^{2}\left(a_{ik}t + b_{ik}\right)
\end{equation}
where $0 \leq w_{k} \leq 1$ and $\sum_{k = 1}^{K} w_{k} = 1$. The factor $2$ is added to simplify subsequent derivations. Our initial intuition is that the Taylor expansion of a sinusoidal function at an arbitrary point is a polynomial of \textit{infinite} degrees, even though the coefficients are constrained to fit the sinusoidal function. We expect this formulation to have comparable expressive power with polynomials of \textit{finite} degrees but with free coefficients as employed in SOS~\citep{pmlr-v97-jaini19a}. The convex sum allows us to bundle different sinusoidal functions to increase expressive power, while keeping the overall $f$ still a sinusoidal function. 

The real benefit of replacing polynomials of finite degrees with sinusoidals becomes more evident after we solve the integral in Equation~\ref{eqn: integral} analytically, assuming $a_{ik} \neq 0$:
\begin{align}
    z^{l}_{i} 
    &= \int_{0}^{z^{l - 1}_{i}} f(t;\matr{h}_{i}) dt + d_{i} \nonumber \\
    &= \sum_{k = 1}^{K} w_{k} \int_{0}^{z^{l - 1}_{i}} 2\sin^{2}\left(a_{ik}t + b_{ik}\right)dt + d_{i} \nonumber \\
    &= \sum_{k = 1}^{K} w_{k} \int_{0}^{z^{l - 1}_{i}} 1 - \cos\left[2\left(a_{ik}t + b_{ik}\right)\right]dt + d_{i} \nonumber \\
    &=  z^{l - 1}_{i} \underbrace{- \sum_{k = 1}^{K} \frac{w_{k}}{2a_{ik}} \sin(2a_{ik}\cdot z^{l - 1}_{i} + 2b_{ik}) + 
    \overbrace{\sum_{k = 1}^{K} \frac{w_{k}}{2a_{ik}}\sin\left(2b_{ik}\right) + d_{i}}^{\text{Constant}}}_{\text{Residual}}
    \label{eqn: sin_transform}
\end{align}
The resultant transformation bears a close resemblance to the \textit{contractive residual transformation} described in Section~\ref{sec: residual} and in fact, it is equivalent to a single-layer feed-forward network with a residual connection and a sine activation function. A small technicality we need to address is the assumption of $a_{ik} \neq 0$. In practice, it is hard to exclude only a single value from a rather continuous weight vector; however, we are allowed to constrain $a_{ik} > 0$ (e.g., using a softplus) \textit{without} hindering the expressive power of our proposed transformer, because sine is an odd function:
\begin{equation}
    \frac{\sin\left[2(-a_{ik})z^{l - 1}_{i} + 2b_{ik}\right]}{2(-a_{ik})} = \frac{\sin\left[2(a_{ik})z^{l - 1}_{i} - 2b_{ik}\right]}{2(a_{ik})}
\end{equation}
So a negation of $a_{ik}$ can be absorbed by a negation of $b_{ik}$. Moreover, we could have used a cosine function or a mixture of sines and cosines as the sinusoidal function, but all these subtle differences can be absorbed by the bias term $b_{ik}$. 

The intimate connection between our sinusoidal transformer specified in Equation~\ref{eqn: sin_transform} and the contractive residual transformation from Section~\ref{sec: residual} makes the fast inversion procedure described in Section~\ref{sec: residual} applicable to our proposed transformer. However, we need to ensure the residual term is always a contraction mapping. Note that, if the parameters $\left[\matr{h}_{i}, d_{i}\right]$ are produced by an independent conditioner, the residual term is already a Lipschitz-continuous function with a Lipschitz constant $M \leq 1$, since its first derivative is bounded above by $1$ in absolute value. Inspired by the work ReZero~\citep{bachlechner2020rezero}, we introduce a new parameter $-1 < \alpha_{i} < 1$ called the \textit{residual weight} and re-write the residual term as
\begin{equation}\label{eqn: residual_fun}
    g(z^{l - 1}_{i}) = -\alpha_{i} \sum_{k = 1}^{K} \frac{w_{k}}{2a_{ik}} \sin(2a_{ik}\cdot z^{l - 1}_{i} + 2b_{ik}) + \sum_{k = 1}^{K} \frac{w_{k}}{2a_{ik}}\sin\left(2b_{ik}\right) + d_{i}
\end{equation}
The residual weight $\alpha_{i}$ is only applied to the term associated with the input $z^{l - 1}_{i}$. In practice, such a parameter can be easily introduced by applying the $\tanh$ function to an unconstrained weight. Now, we must have $\abs{g^{\prime}(z^{l - 1}_{i})} < 1$ and therefore $g$ is a contraction mapping, \textbf{under the assumption of an independent conditioner}. Consequently, a single sinusoidal transformer $\tau (z_{i}^{l - 1}) = z_{i}^{l - 1} + g(z_{i}^{l - 1})$ is indeed fast invertible. 

\subsection{LDU Blocks}\label{sec: ldu}

An independent conditioner ensures $g$ in Equation~\ref{eqn: residual_fun} is a contraction mapping because it produces input-agnostic parameters for $g$. However, if all component-wise transformers $\tau$ are input-agnostic, each constituent transformation $T_{l}^{-1}$ in Equation~\ref{eqn: decomp} would just amount to a simple scaling transformation. In order to model complex distributions, each $T_{l}^{-1}$ needs to somehow take the input into account and in fact, our $T_{l}^{-1}$ should be fully autoregressive since we aim to formulate our Sinusoidal Flow as a fully autoregressive flow. How can $T_{l}^{-1}$ be fully autoregressive while each component-wise transformer $\tau$ remains input-agnostic? 

One solution stems from the fact that every invertible matrix, such as the Jacobian of $T_{l}^{-1}$, $\matr{J}_{T^{-1}_{l}}$, has a unique ``LDU'' factorisation:
\begin{equation}
    \matr{J}_{T^{-1}_{l}} = \matr{L}_{T^{-1}_{l}} \matr{D}_{T^{-1}_{l}} \matr{U}_{T^{-1}_{l}}
\end{equation}
where $\matr{L}_{T^{-1}_{l}}$ is a lower unitriangular matrix, $\matr{D}_{T^{-1}_{l}}$ is a diagonal matrix and $\matr{U}_{T^{-1}_{l}}$ is an upper unitriangular matrix. This factorisation suggests that the overall transformation $T_{l}^{-1}$ is equivalent to first applying a transformation $U_{l}^{-1}$ whose Jacobian is \textbf{upper unitriangular}, and then applying a transformation $D_{l}^{-1}$ with a \textbf{diagonal} Jacobian, followed by another transformation $L_{l}^{-1}$ whose Jacobian is \textbf{lower unitriangular}:
\begin{equation}
    T_{l}^{-1} = L_{l}^{-1} \circ D_{l}^{-1} \circ U_{l}^{-1}
\end{equation}

This way, the Jacobian of the overall transformation $T_{l}^{-1}$ is a dense matrix in gene\-ral, similar to that of a contractive residual transformation described in Section~\ref{sec: residual}, but its determinant is nevertheless trivial to compute, since $\mathrm{det}\, \matr{J}_{T^{-1}_{l}} = \mathrm{det}\, \matr{D}_{T^{-1}_{l}}$, which just evaluates to a product of the diagonal entries. Furthermore, it is not so hard to infer from their respective Jacobians what kind of transformations $L_{l}^{-1}$, $D_{l}^{-1}$ and $U_{l}^{-1}$ should be. 

\begin{table*}[t]
\caption{Key properties of the $L_{l}^{-1}$, $D_{l}^{-1}$ and $U_{l}^{-1}$ transformations and their Jacobians. The transformation $D_{l}^{-1}$ employs an independent conditioner. }
\vskip -0.15in
\label{tbl: ldu}
\begin{center}
\begin{sc}
\scalebox{0.9}{
\begin{tabular}{l|c|c|c}
\toprule
Transformation & $L_{l}^{-1}$ & $D_{l}^{-1}$ & $U_{l}^{-1}$ \\
\midrule
Conditioner    
& $d_{i} = c_{i}(\matr{z}_{<i}^{l - 1})$ 
& $\left[\matr{h}_{i}, d_{i}\right] = c_{i}(\cdot)$ 
& $d_{i} = c_{i}(\matr{z}_{>i}^{l - 1})$      \\
\midrule
Transformer 
& $z^{l}_{i} = z^{l - 1}_{i} + d_{i}$ 
& $z^{l}_{i} = z^{l - 1}_{i} + g(z^{l - 1}_{i} ; \matr{h}_{i}, d_{i})$ 
& $z^{l}_{i} = z^{l - 1}_{i} + d_{i}$ \\
\midrule
Jacobian & $\mathrm{Lower \, unitriangular}$ & $\mathrm{Diagonal}$ & $\mathrm{Upper \, unitriangular}$ \\
\midrule
Jacobian Det. &  1 & $\prod_{i} [1 + g^{\prime}(z^{l - 1}_{i})]$ & 1 \\
\bottomrule
\end{tabular}
}
\end{sc}
\end{center}
\vskip -0.2in
\end{table*}

\begin{figure}[b]
\vskip -0.1in
\begin{center}
\centerline{
\includegraphics[width=\textwidth]{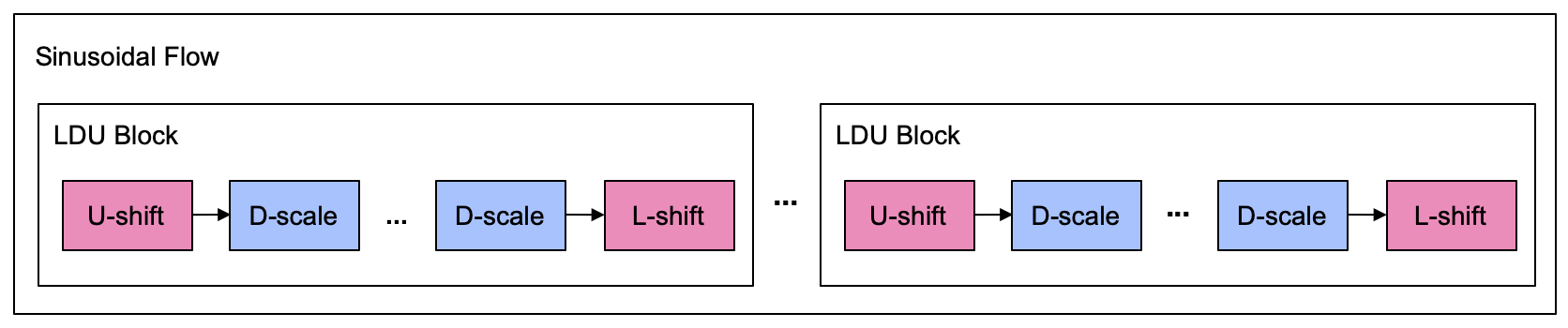}}
\vskip -0.1in
\caption{Our proposed Sinusoidal Flow consists of many LDU blocks as the constituent transformations. Each LDU block encompasses a number of diagonal scaling transformations interleaved with two shift transformations. }
\label{fig: ldu}
\end{center}
\vskip -0.4in
\end{figure}

As summarised in Table~\ref{tbl: ldu}, $L_{l}^{-1}$ and $U_{l}^{-1}$ represent a ``shift'' transformation whose amount of shift to each input component is determined autoregressively on either preceding or succeeding input components. The diagonal scaling transformation $D_{l}^{-1}$ applies to each input component (a chain of) invertible sinusoidal transformers, each paired with an independent conditioner. As Figure~\ref{fig: ldu} shows, by implementing each constituent transformation $T_{l}^{-1}$ as an ``LDU block'' that encompasses the three transformations, we manage to introduce full autoregressiveness into our Sinusoidal Flow while keeping the sinusoidal transformers still fast inverti\-ble. It also suggests a new paradigm in designing autoregressive flows: instead of leveraging a conditioner to compute the parameters of the core transformations, which could result in a large number of trainable parameters, interleave \textit{unconditional} core transformations with conditional shift transformations. As we shall see in Section~\ref{sec: real_world}, our Sinusoidal Flow achieves comparable performance despite following this new paradigm.  

The LDU blocks naturally make the overall determinant easy to compute. Therefore, in the following two sections, we scrutinise our Sinusoidal Flow through the lens of the other two desirable properties we put forward, in order to develop deeper insights into it. 

\subsection{Expressiveness}\label{sec: expressiveness}

While the shift transformations bring in full autoregressiveness, the ultimate source of expressiveness lies in the core ``D-scale'' transformations. Therefore, we first investigate the expressiveness of our proposed sinusoidal transformer that serves as ``D-scale" in Figure~\ref{fig: ldu}. 


\renewcommand{\thefootnote}{\fnsymbol{footnote}}

\begin{wrapfigure}{r}{0.45\textwidth} 
    \centering
    \includegraphics[width=0.9\linewidth]{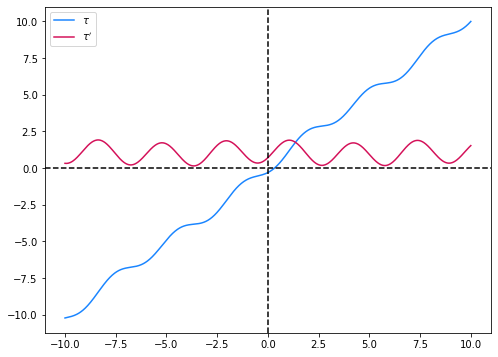}
    \caption{A plot of an arbitrary sinusoidal transformer $\tau$\protect\footnotemark 
    and its first derivative $\tau^{\prime}$. $\tau$ resembles a staircase running along the diagonal with infinitely many inflection points. }
\label{fig: sinusoidal}
\vskip -0.4in
\end{wrapfigure}

Figure~\ref{fig: sinusoidal} shows a plot of a sinusoidal transformer $\tau$ with arbitrary parameter values. We can observe that it assumes a ``staircase'' shape along the diagonal line $y = x$ with an infinite number of inflection points. As noted in~\citet{pmlr-v80-huang18d,pmlr-v97-jaini19a}, these inflection points are actually critical to modelling multi-modal distributions, especially when the modes are well separated. If in Equation~\ref{eqn: formulation} we take $p_{\matr{x}}$ to be a uniform mixture of well-separated one-dimensional Gaussians and $p_{\matr{z}}$ to be a simple one-dimensional standard Gaussian, then the transformation $T^{-1}$ that takes any $x$ into $z$ must
\begin{itemize}
    \item plateau in regions between any two modes of $p_{\matr{x}}$ where there is nearly zero density; and
    \item admit a sharp jump around a mode of $p_{\matr{x}}$.
\end{itemize}
This can be seen directly from Equation~\ref{eqn: formulation}. When $p_{\matr{x}}\left(x\right)$ is nearly zero, either $p_{\matr{z}}\left(T^{-1}\left(x\right)\right)$ or $\abs{\mathrm{det}\, \matr{J}_{T^{-1}}\left(x\right)}$ must be close to zero; however, being a standard Gaussian, $p_{\matr{z}}$ is near zero only at its boundaries, so for the $99.7\%$ of $x$ that gets mapped to $[-3, 3]$, $\abs{\mathrm{det}\, \matr{J}_{T^{-1}}\left(x\right)} = \abs{dT^{-1} / dx}$ must be close to zero and hence $T^{-1}$ plateaus. On the other hand, when $p_{\matr{x}}$ undergoes a drastic change around a mode, $\abs{dT^{-1} / dx}$ must first explode and then vanish, forcing $T^{-1}$ to make a sharp jump, as $p_{\matr{z}}$ is not flexible enough to change so quickly. Therefore, accurately modelling a multi-modal distribution with well-separated modes potentially requires a $T^{-1}$ capable of creating a large number of sharp jumps like a step function. 

\footnotetext{$\tau\left(x\right) = x - 0.8\left[0.5\sin(2x+1) + 0.3\sin(-0.1x) + 0.2\sin(0.7x - 3)\right]$}

\begin{figure}[ht]
\vskip -0.2in
\begin{center}
\centerline{
\includegraphics[width=\textwidth]{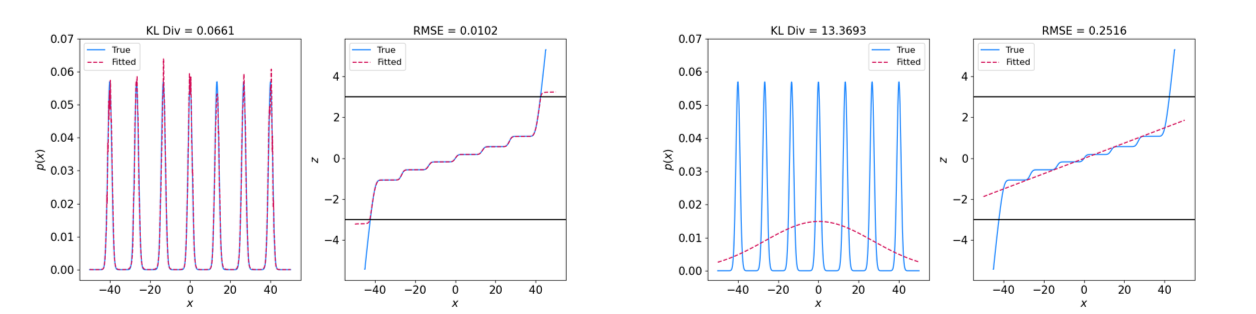}}
\vskip -0.15in
\caption{Comparison of Sinusoidal Flow and MAF~\citep{NIPS2017_6c1da886} on fitting a multi-modal distribution. \textbf{Left}: density and transformation learned by Sinusoidal Flow. \textbf{Right}: density and transformation learned by MAF. With sinusoidal transformers, Sinusoidal Flow is able to learn the step-function-like true transformation while MAF can only learn an affine transformation. } 
\label{fig: 7gaussians}
\end{center}
\vskip -0.2in
\end{figure}

We claim that our proposed sinusoidal transformer has this inductive bias built in, which makes it suitable for modelling multi-modal distributions. Indeed, as shown in Figure~\ref{fig: sinusoidal}, it already resembles a step function at initialisation and the distance between any two jumps can be easily adjusted to fit the data by adjusting the periods of the constituent sinusoidals. To further verify our claim, we fit an unconditional Sinusoidal Flow composed of purely ``D-scale'' transformations to data generated from a uniform mixture of seven one-dimensional Gaussians, and compare the learned density and transformation with that learned by an MAF~\citep{NIPS2017_6c1da886}. We can see from Figure~\ref{fig: 7gaussians} that our Sinusoidal Flow is not only able to learn the true density well, but can also recover the true transformation $T^{-1}$ which is characterised by ``six steps''. Interestingly, the learned transformation only approximates (in fact, only \emph{needs to} approximate) the true transformation well for $z$ in the interval $[-3, 3]$, because outside the interval $p_{\matr{z}}$ is nearly zero.




\subsection{Fast Inversion}
We now discuss the invertibility of the transformations $L_{l}^{-1}$, $D_{l}^{-1}$ and $U_{l}^{-1}$, as they jointly determine the invertibility of each LDU block. With sinusoidal transformers acting on each input component, we can write $D_{l}^{-1}$ in vector form as
\begin{equation}\label{eqn: vector_form}
    \matr{z}^{l} = D_{l}^{-1}(\matr{z}^{l - 1}) = \matr{z}^{l - 1} + G(\matr{z}^{l - 1})
\end{equation}
where $G: \mathbb{R}^{D} \mapsto \mathbb{R}^{D}$ applies the residual function $g$ defined in Equation~\ref{eqn: residual_fun} to each component of the input $\matr{z}^{l - 1}$. We have shown that the scalar function $g$ is a contraction mapping under the assumption of an independent conditioner, but not yet that the vector-valued function $G$ is also a contraction mapping, which we now prove. 
\begin{proposition}\label{prop: contraction}
The function $G$ defined in Equation~\ref{eqn: vector_form} is a contraction mapping. Specifically, for any $\matr{u}, \matr{v} \in \mathbb{R}^{D}$, $\norm{G\left(\matr{u}\right) - G\left(\matr{v}\right)} \leq M \norm{\matr{u} - \matr{v}}$ for some $0 \leq M < 1$. 
\end{proposition}
\begin{proof}
It is sufficient to prove the proposition with respect to the $\ell_{1}$ norm, since all norms are equivalent in $\mathbb{R}^{D}$. 

We have $\norm{G\left(\matr{u}\right) - G\left(\matr{v}\right)} = \sum_{i} \abs{g\left(u_{i}\right) - g\left(v_{i}\right)}$ and $\norm{\matr{u} - \matr{v}} = \sum_{i} \abs{u_{i} - v_{i}}$ by definition. Because $g$ is a contraction mapping, there exists an $0 \leq M_{i} < 1$ for each $i$ such that $\abs{g\left(u_{i}\right) - g\left(v_{i}\right)} \leq M_{i} \abs{u_{i} - v_{i}}$. Let $0 \leq M = \max_{i} M_{i} < 1$, then we must have
\begin{equation*}
    \sum_{i} \abs{g\left(u_{i}\right) - g\left(v_{i}\right)} \leq \sum_{i} M_{i}\abs{u_{i} - v_{i}} \leq M \sum_{i} \abs{u_{i} - v_{i}}
\end{equation*}
which is the same as $\norm{G\left(\matr{u}\right) - G\left(\matr{v}\right)} \leq M \norm{\matr{u} - \matr{v}}$. 
\end{proof}
\vskip -0.1in
\begin{wrapfigure}{r}{0.5\textwidth}
\vskip -0.32in
    \begin{minipage}{0.5\textwidth}
      \begin{algorithm}[H]
\SetAlgoLined
   \caption{Inversion of $D_{l}^{-1}$}
   \label{alg: inversion}
\begin{algorithmic}
   \STATE {\bfseries Input:} output $\matr{z}^{l}$ from $D_{l}^{-1}$; max number of iterations $n$; error tolerance $\epsilon$
   \STATE Initialise $\matr{z}^{l}_{0} \defeq \matr{z}^{l}$, $j \defeq -1$.
   \REPEAT
   \STATE $j \defeq j + 1$
   \STATE $\matr{z}^{l}_{j+1} \defeq \matr{z}^{l} - (D_{l}^{-1}(\matr{z}^{l}_{j}) - \matr{z}^{l}_{j})$
   \UNTIL{$\lVert\matr{z}^{l}_{j+1} - \matr{z}^{l}_{j}\rVert < \epsilon$} or $j == n$
   \STATE \Return{$\matr{z}^{l}_{j}$}
\end{algorithmic}
\end{algorithm}
    \end{minipage}
\vskip -0.5in
  \end{wrapfigure}

Proposition~\ref{prop: contraction} formally establishes that $D_{l}^{-1}$ is a contractive residual transformation as described in Section~\ref{sec: residual}, and as a result, $D_{l}^{-1}$ is fast invertible via Algorithm~\ref{alg: inversion} adapted from~\citet{pmlr-v97-behrmann19a}. In addition, the Banach fixed-point theorem guarantees Algorithm~\ref{alg: inversion} converges with \textit{any} starting points $\matr{z}_{0}^{l}$ at a rate \textit{exponential} in the number of iterations~\citep{pmlr-v97-behrmann19a}.

It may be attempting to draw the same conclusion for $L_{l}^{-1}$ and $U_{l}^{-1}$ since they both employ a similar residual structure, but there are no strong theoretical guarantees for them because their conditioners produce \textit{input-dependent} shifts that dynamically determine the actual transformations. While we could have used spectral normalisation~\citep{miyato2018spectral} to regularise their conditioners, we observe in practice that Algorithm~\ref{alg: inversion} is still app\-licable if we use a large number of LDU blocks to share the overall complexity, which is possible only because each LDU block has a built-in residual connection. In the (unlikely) worst case, Algorithm~\ref{alg: inversion} would just resort to a fully vectorised sequential inversion.

Leveraging LDU blocks to introduce full autoregressiveness, we manage to guarantee the core ``D-scale'' transformations are fast invertible, while the auxiliary shift transformations remain empirically fast invertible. As a result, our Sinusoidal Flow is among the few fully autoregressive flows known to be fast invertible.~\citet{NEURIPS2019_70a32110} proposes a Newton-like fixed-point iteration to invert fully autoregressive flows, but their algorithm is only locally convergent and sensitive to starting points; RQ-NSF~\citep{NEURIPS2019_7ac71d43} offers an analytic inverse but their fully autoregressive variant still requires a sequential inversion.




\section{Experiments}\label{sec: experiments}

In this section, we evaluate our Sinusoidal Flow against some common benchmarks to further assess its expressiveness and fast inversion. 

\subsection{Toy Datasets}
\begin{figure*}[hb]
\vskip 0.1in
\begin{center}
\centerline{
\includegraphics[width=0.62\textwidth]{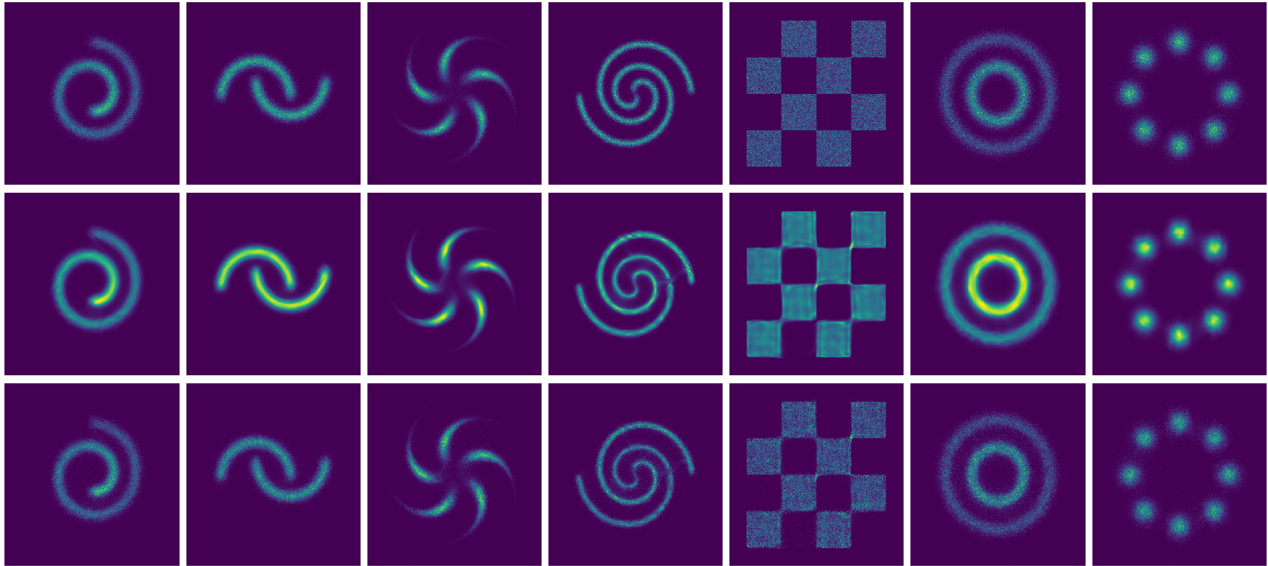}}
\caption{A Sinusoidal Flow composed of 16 LDU blocks fit to a set of toy density estimation problems proposed by~\citet{grathwohl2018scalable}. \textbf{Top row}: the true distributions. \textbf{Middle row}: the learned distributions. \textbf{Bottom row}: samples generated from the learned distributions. }
\label{fig: shapes}
\end{center}
\vskip -0.2in
\end{figure*}

Following~\citet{NEURIPS2019_2a084e55}, we first apply our Sinusoidal Flow to a set of toy density estimation problems proposed by~\citet{grathwohl2018scalable}. As shown in Figure~\ref{fig: shapes}, our Sinusoidal Flow is not only able to capture multi-modal or even discontinuous distributions well, but can also be reliably inverted to generate realistic-looking samples from the learned distributions using Algorithm~\ref{alg: inversion}, which further corroborates our claim in Section~\ref{sec: expressiveness} that the inductive bias built into our sinusoidal transformers, which favours step-function-like transformations, enables Sinusoidal Flow to accurately model multi-modal distributions. 

\begin{table*}[ht]
\vskip -0.3in
\caption{Average negative test log-likelihood in nats over five runs (except for RQ-NSF) on the five real-world datasets from~\citet{NIPS2017_6c1da886}.  Lower is better.}
\vskip -0.35in
\label{tbl: uci_results}
\begin{center}
\begin{sc}
\scalebox{0.65}{
\begin{tabular}{lccccc}
\toprule
Method & POWER & GAS & HEPMASS & MINIBOONE & BSDS300  \\
\midrule
FFJORD~\citep{grathwohl2018scalable} & -0.46 $\pm$ 0.01 & -8.59 $\pm$ 0.12 & 14.92 $\pm$ 0.08 & 10.43 $\pm$ 0.04 & -157.40 $\pm$ 0.19 \\
\midrule
RealNVP~\citep{DBLP:conf/iclr/DinhSB17}    & -0.17 $\pm$ 0.01 & -8.33 $\pm$ 0.14 & 18.71 $\pm$ 0.02 & 13.55 $\pm$ 0.49 & -153.28 $\pm$ 1.78 \\
Glow~\citep{NEURIPS2018_d139db6a} & -0.17 $\pm$ 0.01& -8.15 $\pm$ 0.40 & 19.92 $\pm$ 0.08 & 11.35 $\pm$ 0.07 & -155.07 $\pm$ 0.03 \\
RQ-NSF(C)~\citep{NEURIPS2019_7ac71d43} & -0.64 $\pm$ 0.01 & \textbf{-13.09} $\pm$ 0.02 & 14.75 $\pm$ 0.03 & 9.67 $\pm$ 0.47 & -157.54 $\pm$ 0.28 \\
\midrule
MAF~\citep{NIPS2017_6c1da886}     & -0.24 $\pm$ 0.01 & -10.08 $\pm$ 0.02 & 17.70 $\pm$ 0.02 & 11.75 $\pm$ 0.44 & -155.69 $\pm$ 0.28  \\
TAN~\citep{pmlr-v80-oliva18a}      & -0.60 $\pm$ 0.01 & -12.06 $\pm$ 0.02 & \textbf{13.78} $\pm$ 0.02 & 11.01 $\pm$ 0.48 & \textbf{-159.80} $\pm$ 0.07 \\
NAF~\citep{pmlr-v80-huang18d}      & -0.62 $\pm$ 0.01 & -11.96 $\pm$ 0.33 & 15.09 $\pm$ 0.40 & \textbf{8.86} $\pm$ 0.15 & -157.73 $\pm$ 0.30    \\
B-NAF~\citep{pmlr-v115-de-cao20a}   & -0.61 $\pm$ 0.01 & -12.06 $\pm$ 0.09 & 14.71 $\pm$ 0.38 & 8.95 $\pm$ 0.07 & -157.36 $\pm$ 0.03 \\
SOS~\citep{pmlr-v97-jaini19a}   & -0.60 $\pm$ 0.01 & -11.99 $\pm$ 0.41 & 15.15 $\pm$ 0.10 & 8.90 $\pm$ 0.11 & -157.48 $\pm$ 0.41  \\
UMNN-MAF~\citep{NEURIPS2019_2a084e55}   & -0.63 $\pm$ 0.01 & -10.89 $\pm$ 0.70 & 13.99 $\pm$ 0.21 & 9.67 $\pm$ 0.13 & -157.98 $\pm$ 0.01  \\
RQ-NSF(AR)~\citep{NEURIPS2019_7ac71d43} & \textbf{-0.66} $\pm$ 0.01 & \textbf{-13.09} $\pm$ 0.02 & 14.01 $\pm$ 0.03 & 9.22 $\pm$ 0.48 & -157.31 $\pm$ 0.28 \\
\midrule
\textbf{Sinusoidal Flow (Ours)}   & -0.59 $\pm$ 0.00 & -12.15 $\pm$ 0.05 & 16.10 $\pm$ 0.06 & 9.25 $\pm$ 0.03 & -156.75 $\pm$ 0.03 \\
\bottomrule
\end{tabular}
}
\end{sc}
\end{center}
\vskip -0.2in
\end{table*}


\subsection{Real-world Datasets}\label{sec: real_world}

We compare our Sinusoidal Flow with several other normalising flows on five high-dimensional real-world datasets (POWER, GAS, HEPMASS, MINIBOONE and BSDS300) proposed by~\citet{NIPS2017_6c1da886}. From Table~\ref{tbl: uci_results} we observe that, while no method can claim the best performance across all five datasets, Sinusoidal Flow is comparable to those top-performing methods on each dataset, with an (additional) advantage of being fast invertible. 

\begin{wraptable}{r}{0.45\textwidth}
\vskip -0.2in
\caption{{Bits/dim on MNIST and CIFAR-10. Lower is better.}}
\label{tbl: mnist_cifar10}
\centering
\begin{sc}
\scalebox{0.54}{
\begin{tabular}{lcc}
\toprule
Method & MNIST & CIFAR-10 \\
\midrule
FFJORD~\citep{grathwohl2018scalable} & \textbf{0.99} & 3.40  \\
i-ResNet~\citep{pmlr-v97-behrmann19a} & 1.06 & 3.45 \\
\midrule
RealNVP~\citep{DBLP:conf/iclr/DinhSB17}    & 1.06 & 3.49  \\
Glow~\citep{NEURIPS2018_d139db6a} & 1.05 & \textbf{3.35} \\
RQ-NSF(C)~\citep{NEURIPS2019_7ac71d43} & - & 3.38 \\
\midrule
MAF~\citep{NIPS2017_6c1da886}     & 1.89 & 4.31  \\
TAN~\citep{pmlr-v80-oliva18a}      & 1.19 & 3.98 \\
SOS~\citep{pmlr-v97-jaini19a}   & 1.81 & -  \\
UMNN-MAF~\citep{NEURIPS2019_2a084e55}   & 1.13 & -  \\
\midrule
\textbf{Sinusoidal Flow (Ours)}   & 1.06 & 3.51 \\
\bottomrule
\end{tabular}
}
\end{sc}
\vskip -0.2in
\end{wraptable} 

\begin{figure*}[ht]
\vskip -0.2in
\begin{center}
\centerline{
\includegraphics[width=0.8\textwidth]{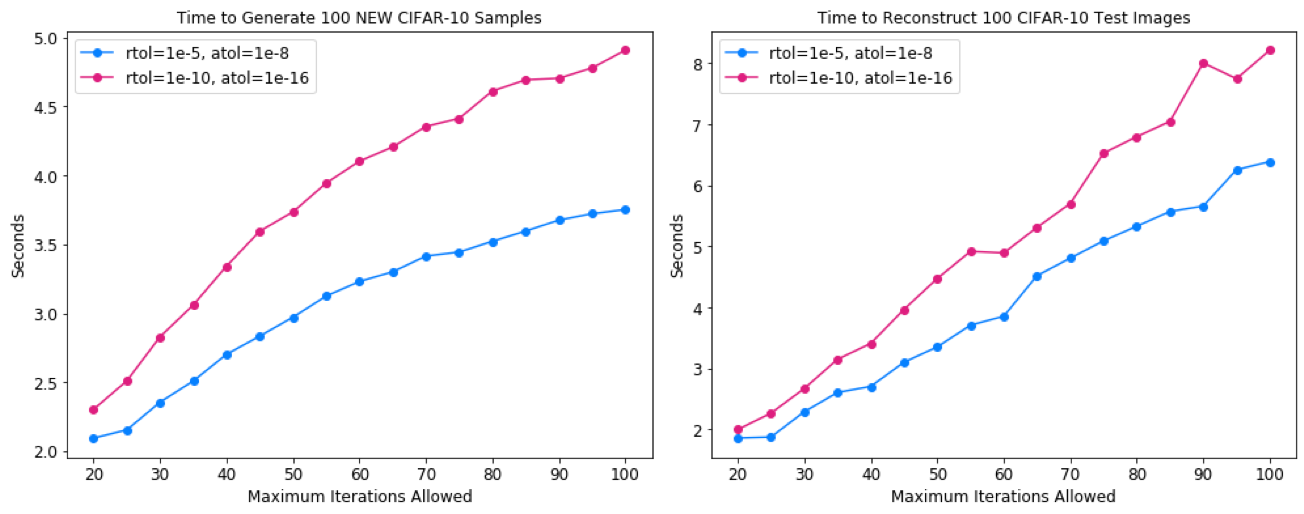}}
\vskip -0.1in
\caption{\textbf{Left}: Runtimes for generation. \textbf{Right}: Runtimes for reconstruction. \texttt{rtol} and \texttt{atol} refer to the corresponding arguments of a typical \texttt{allclose} function.}
\vskip -0.2in
\label{fig: mnist_cifar10}
\end{center}
\end{figure*}

To further assess its efficacy in model\-ling high-dimensional data like images, we apply our Sinusoidal Flow to MNIST~\citep{726791} and CIFAR-10~\citep{Krizhevsky09learningmultiple} pre-processed by~\citet{NIPS2017_6c1da886}. Unlike existing work that uses a carefully engineered multi-scale convolutional architecture~\citep{DBLP:conf/iclr/DinhSB17}, we simply adopt a miniature Pixel\-CNN~\citep{NIPS2016_b1301141} as the conditioners for the shift transformations in each LDU block. As shown in Table~\ref{tbl: mnist_cifar10}, compared to other fully autoregressive flows that ever managed to model images without an explosion in parameters, our Sinusoidal Flow achieves results much closer to that achieved by methods with convolutional architectures. Moreover, in Figure~\ref{fig: mnist_cifar10} we can see some reasonable runtimes for both generating and reconstructing CIFAR-10 images at different tolerance levels and maximum iterations allowed at each LDU block, which reaffirms that our Sinusoidal Flow is fast invertible. 


\section{Conclusions}
We recognise expressiveness, fast inversion and exact Jacobian determinant as three proper\-ties desirable for normalising flows. Leveraging the fact that the integral of a sinusoidal function squared resembles a residual transformation, Sinusoidal Flow represents an interesting attempt at integrating the respective strengths of fully autoregressive flows and contractive residual flows in order to excel at all three dimensions. Our unique design of the LDU blocks offers a new way of introducing full autoregressiveness without hurting the invertibility of the core transformations and naturally ensures an easy Jacobian determinant. 





\bibliography{ms}

\end{document}


\maketitle
\appendix
\section{Code}
Our code is available at \url{https://github.com/weiyumou/ldu-flow}.

\section{Experiment Details}

In this section, we describe the architectural and training-specific details about our Sinusoidal Flow applied to the two-dimensional toy dataset from~\citet{grathwohl2018scalable}, the five high-dimensional tabular datasets proposed by~\citet{NIPS2017_6c1da886} and a version of MNIST~\citep{726791} and CIFAR-10~\citep{Krizhevsky09learningmultiple} pre-processed by~\citet{NIPS2017_6c1da886}.  In addition, please refer to our code for more details. 

\subsection{Toy and Tabular Datasets}

Consistent with existing work, we use the masked linear layers from MADE~\citep{pmlr-v37-germain15} as the conditioners for the shift transformations. Table~\ref{tbl: toy_tabular} summarises the experiment details. We use the Adam optimiser~\citep{DBLP:journals/corr/KingmaB14}, and either exponential decay or cosine annealing~\citep{DBLP:journals/corr/LoshchilovH16a} for reducing learning rates over time. 

\begin{table*}[hb]
\caption{Architectural and training-specific details about our Sinusoidal Flow applied to the toy and tabular datasets. ``Embedding dim'' refers to the number of parallel sinusoidal functions bundled in the convex sum inside a sinusoidal transformer, denoted as $K$ in the main text. }
\vskip -0.15in
\label{tbl: toy_tabular}
\begin{center}
\begin{sc}
\scalebox{0.65}{
\begin{tabular}{lcccccc}
\toprule
Data Set & 2D TOY & POWER & GAS & HEPMASS & MINIBOONE & BSDS300  \\
\midrule
\textbf{Architectural} \\
\hspace{1em} \# LDU blocks & 16 & 12 & 12 & 12 & 12 & 12 \\
\hspace{1em} \# D-scale per block & 4 & 4 & 4 & 4 & 4 & 4 \\
\hspace{1em} Embedding dim ($K$) & 4 & 4 & 4 & 4 & 4 & 4 \\
\hspace{1em} Hidden size & 100 & 256-256 & 256-256 & 512-512 & 256-256 & 512-512 \\
\hspace{1em} Dropout & - & - & - & - & 0.3 & 0.1 \\
\midrule
\textbf{Training-specific} \\
\hspace{1em} \# Steps   & 50K & 1.2M & 2M & 1M & 125K & 400K \\
\hspace{1em} Batch size   & 128 & 512 & 128 & 128 & 128 & 512 \\
\hspace{1em} Learning rate   & \num{1e-3} & \num{5e-4} & \num{1e-3} & \num{1e-3} & \num{5e-4} & \num{5e-4} \\
\hspace{1em} LR decay & - & \text{cosine} & 0.99 & 0.99 & \text{cosine} & \text{cosine} \\
\hspace{1em} Weight decay   & - & - & \num{1e-5} & \num{5e-4} & \num{1e-3} & - \\
\bottomrule
\end{tabular}
}
\end{sc}
\end{center}
\vskip -0.2in
\end{table*}

\subsection{MNIST and CIFAR-10}

For modelling image data, we use a miniature Pixel\-CNN~\citep{NIPS2016_b1301141} as the conditioners for the shift transformations. The experiment details are shown in Table~\ref{tbl: mnist_cifar10}. The AdamW optimiser~\citep{loshchilov2018decoupled} appears to help stabilise training for CIFAR-10. We also reduce the learning rate by a factor of $0.9$ whenever the validation loss stops improving for two epochs. 

\begin{table*}[hb]
\caption{Architectural and training-specific details about our Sinusoidal Flow applied to MNIST and CIFAR-10. }
\vskip -0.15in
\label{tbl: mnist_cifar10}
\begin{center}
\begin{sc}
\scalebox{0.7}{
\begin{tabular}{lcc}
\toprule
Data Set & MNIST & CIFAR-10   \\
\midrule
\textbf{Architectural} \\
\hspace{1em} \# LDU blocks & 12 & 12  \\
\hspace{1em} \# D-scale per block & 4 & 4  \\
\hspace{1em} Embedding dim ($K$) & 4 & 4  \\
\hspace{1em} \# Feature maps & 16 & 16 \\
\hspace{1em} \# Residual blocks & 2 & 2 \\
\midrule
\textbf{Training-specific} \\
\hspace{1em} \# Steps   & 200K & 140K  \\
\hspace{1em} Batch size  & 128 & 128  \\
\hspace{1em} Learning rate   & \num{1e-3} & \num{5e-4}  \\
\hspace{1em} LR decay & \text{cosine} & 0.9 \\
\hspace{1em} Weight decay  & \num{1e-2} & \num{1e-1} \\
\bottomrule
\end{tabular}
}
\end{sc}
\end{center}
\vskip -0.2in
\end{table*}

\section{Reconstruction Analysis}

\begin{figure*}[ht]
\begin{center}
\centerline{
\includegraphics[width=0.9\textwidth]{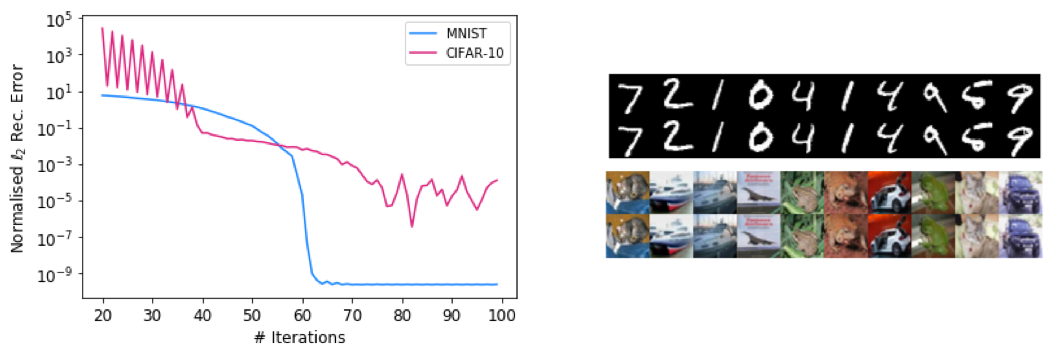}}
\vskip -0.1in
\caption{\textbf{Left}: $\ell_{2}$ reconstruction error versus maximum number of iterations allowed. \textbf{Right}:  original images (odd rows) versus their reconstructions (even rows). }
\vskip -0.2in
\label{fig: mnist_cifar10}
\end{center}
\end{figure*}

To develop further insights into the fast invertibility of our Sinusoidal Flow, we perform an analysis on how well it can reconstruct MNIST and CIFAR-10 images from the mapped $\matr{z}$ of some input images. A visually indiscernible reconstruction would require an accurate inversion of our Sinusoidal Flow. The left of Figure~\ref{fig: mnist_cifar10} shows that the reconstruction errors for both datasets decrease fairly quickly as more iterations are allowed for inversion at each LDU block. And in fact, with as few as 70 iterations allowed, we are able to obtain visually indiscernible reconstructions for both datasets as shown on the right of Figure~\ref{fig: mnist_cifar10}.

\bibliography{supplement}